\documentclass{article}
\usepackage{graphicx}
\usepackage{verbatim}
\usepackage[export]{adjustbox}
\usepackage{caption}
\usepackage{subcaption}

\usepackage{PRIMEarxiv}

\usepackage[utf8]{inputenc} % allow utf-8 input
\usepackage[T1]{fontenc}    % use 8-bit T1 fonts
\usepackage{hyperref}       % hyperlinks
\usepackage{url}            % simple URL typesetting
\usepackage{booktabs}       % professional-quality tables
\usepackage{amsfonts}       % blackboard math symbols
\usepackage{nicefrac}       % compact symbols for 1/2, etc.
\usepackage{microtype}      % microtypography
\usepackage{lipsum}
\usepackage{fancyhdr}       % header
\usepackage{graphicx}       % graphics
\graphicspath{{media/}}     % organize your images and other figures under media/ folder

\title{National Origin Discrimination in Deep-learning-powered Automated Resume Screening
}

\author{
  Sihang Li \\
  Santa Clara University \\
  \texttt{sli13@scu.edu}  \\
   \And
  Kuangzheng Li \\
  hireEZ \\
  \texttt{kuangzhengli@hireez.com} \\
  \AND
  Haibing Lu \\
  Santa Clara University \\
  \texttt{hlu@scu.edu} \\
}

\begin{document}
\maketitle

\begin{abstract}
Many companies and organizations have started to use some form of AI-enabled automated tools to assist in their hiring process, e.g. screening resumes, interviewing candidates, performance evaluation. While those AI tools have greatly improved human resource operations efficiency and  provided conveniences to job seekers as well, there are increasing concerns on unfair treatment to candidates, caused by underlying bias in AI systems. Laws around equal opportunity and fairness, like GDPR, CCPA, are introduced or under development, in attempt to regulate AI. However, it is difficult to implement AI regulations in practice, as technologies are constantly advancing and the risk pertinent to their applications can fail to be recognized. This study examined deep learning methods, a recent technology breakthrough, with focus on their application to automated resume screening. One impressive performance of deep learning methods is the representation of individual words as low-dimensional numerical vectors, called word embedding, which are learned from aggregated global word-word co-occurrence statistics from a corpus, like Wikipedia or Google news. The resulting word representations possess interesting linear substructures of the word vector space and have been widely used in downstream tasks, like resume screening. However,  word embedding inherits and reinforces the stereotyping from the training corpus, as deep learning models essentially learn a probability distribution of words and their relations from history data. Our study finds out that if we rely on such deep-learning-powered automated resume screening tools, it may lead to decisions favoring or disfavoring certain demographic groups and raise ethical, even legal, concerns. To address the issue, we developed bias mitigation method. Extensive experiments on real candidate resumes are conducted to validate our study.
\end{abstract}

% keywords can be removed
% \keywords{Human Resource, \And Resume Embedding, \And AI Fairness}

\section{Introduction}
Benefiting from rapid advancements in computer technologies and the availability of big data, AI has reshaped many industries. AI has also been leveraged in talent acquisition and used in recruitment and selection 
\cite{albert2019ai}.  One of the biggest benefits of using AI in recruitment is that it can screen thousands of resumes and shortlist candidates within minutes. The advantage allows recruiters to reduce their manual time and focus on more pertinent matters. AI technologies have been used in different stages of hiring. For example, AI-powered chat bot is used to automate and optimize recruiters' repetitive interactions inside the screening, scheduling, reference checking, etc. Automatic facial expression analysis helps recruiters conduct the initial screening of candidates. AI technologies are also employed and allow recruiters to use games and quizzes to filter candidates. HR technologies indeed provide an effective solution for time-consuming tasks, like resume screening and initial candidate interview, which have been recognized as a bottleneck for organizations to scale and expand their businesses.

However, there are growing concerns about the ethics and lawfulness of the use of AI in hiring \cite{raghavan2020mitigating,van2021machine,dattner2019legal,cowgill2018bias}. In 2018, it was reported that Amazon developed a machine learning-based recruitment program, which was later found to be biased against women. The machine learning algorithm was fed with the company history employment data, where the majority of employees had been men. The imbalanced training data led the system to prefer male candidates over female ones \footnote{https://www.forbes.com/sites/forbeshumanresourcescouncil/2021/10/14/understanding-bias-in-ai-enabled-hiring}. In 2019, EPIC filed a complaint with the Federal Trade Commission against an Human Resource (HR) technology company for its unfair and deceptive practices in violation of the Federal Trade Commission (FTC) Act, e.g. falsely denying the use of facial recognition\footnote{https://epic.org/documents/in-re-hirevue/}.

To address concerns about potential misuse or unintended consequences of AI, however, we have seen many efforts and initiatives on developing technical standards for reliable, robust, and trustworthy AI systems. In addition to that, lawmakers are developing regulations to use police power to ensure good use of AI. At least 17 states in the U.S. in 2022 have introduced general artificial intelligence bills or resolutions. The General Data Protection Regulation (GDPR) is a regulation in European Union law on data protection and privacy, which has policies regarding AI technologies. There are laws around Equal
Opportunity and Fairness. New York City approved a bill, to be effective on January 1, 2023, i.e. ``...a bias audit be conducted on an automated employment decision tool prior to the use of said tool. The bill would also require that candidates or employees that reside in the city be notified about the use of such tools in the assessment or evaluation for hire or promotion, as well as, be notified about the job qualifications and characteristics that will be used by the automated employment decision tool."

However, it is challenging to implement these legal regulations, like the requirement on auditing HR technology, due to many technical reasons. For example, it might be difficult to detect discrimination for certain protected attributes like sexual orientation or disability status, if the data does not contain this information. To better guard against the misuse of AI and its unintended consequence, it is critical to understand how AI systems work and detect/mitigate bias in the algorithms, which requires us to explore how predictive technologies work at each step of the process. 

There are lots of researches on machine learning fairness, responsible AI, etc. This study focuses on the use of AI in HR technology, particularly in the use of natural language processing for resume screening. Resume screening is in large demand in today's job market, especially when applications can be made as simple as one click on some job posting sites, e.g. LinkedIn, Indeed. A resume system typically is responsible for filtering out the resumes that are deemed unqualified or less qualified than the resumes that are selected by the system. Because a hiring process carries ethical responsibility, a resume screening system, which is thought to automate and facilitate the hiring process, should not be exempted from its ethical responsibilities. A resume screening system is expected to be able to select resumes accurately (i.e. the ones selected should match the job requirement) and fairly (i.e. no other factors like gender, nationality, or race, should affect the decision made). However, AI algorithms are trained on existing dataset, which may be not well represented and may contain biases. Such biases will be inherited by the learned AI model and cause disparate impact on the resultant automated decisions \cite{barocas2016big}. Natural Language Processing (NLP) is the technology used to scan textual data in a resume and make candidate recommendations based on the matching between a candidate resume and a job description. The traditional NLP methods represent a document as a bag of words, which can be mathematically represented as a numeral vector with each component indicating the weight of a corresponding word. A popular method to compute the weight is called Term Frequency-Inverse Document Frequency (TF-IDF) \cite{manning2008introduction}. As such, the matching between a candidate resume and a job description can be measured as cosine similarity of their vector representations. The TF-IDF based information retrieval method can be utilized to quickly screen candidate resumes. However, it was reported in
\cite{raghavan2020mitigating} that the method may carry socio-linguistic bias and cause disparate impact on the origin country of the applicants. 
% kan dao zhe
Built upon existing research, this study examines the disparate impact of deep learning (particularly word embedding) based resume screening methods. The disadvantage of TF-IDF is that it is based on the bag-of-words model (the length of a TF-IDF vector is the same size as the vocabulary) and cannot capture information on position, semantics, or co-occurrences of words. Word embedding is a method that represents a word as a short real-valued vector, where each component encodes some characteristic information. Similar words have their representations closer in the vector space. Word embedding is learned by forming a unsupervised learning problem and trained a large corpus. Popular word embeddings include Word2Vec (trained from Google news) \cite{mikolov2013efficient}, GloVe (trained from Wikipedia) \cite{pennington2014glove}. The advantages of word embedding include the small size of the embedding vector and the retaining of the semantic meaning of words and their context information. Due to those advantages, word embedding can be used to support many downstream tasks, including resume screening. However, it was reported in \cite{bolukbasi2016man} that ``word embeddings trained on Google News articles exhibit female/male gender stereotypes to a disturbing extent". The underlying reason is that word embedding essentially captures the association/co-occurrence of words from the training corpus. If the corpus has stereotype information, the pattern will be encoded in the resultant word embedding. This study tries to evaluate the risk of word embedding-based resume screening method, with respect to national origin bias, and propose some mitigation methods. 

\section{Literature Review}
This study is related to multiple research fields, including AI ethics, machine learning fairness, DEI (Diversity, Equality, and Inclusion), NLP, etc. We will review some representative works.

AI ethics concern principles and guidelines that govern the development and implement of AI techniques, with respect to human rights and human dignity. Due to the importance of responsible AI, many institutions and organizations have issued guidance for responsible AI.  In \cite{jobin2019global}, they provided a global landscape of AI ethics guidelines. Although those agreements disagree in some terms, a set of AI principles have received wide consensus:
\begin{itemize}
\item Accountability and Governance \cite{doshi2017accountability} - This principle ensures responsibility for complying with data protection and for demonstrating that compliance in any AI system. It is required to assess and mitigate its risks, and document can demonstrate how the system is compliant and justify the choices made in the process. 
\item Lawfulness, Fairness, and Transparency - AI lawfulness states that the development, deployment, and use of AI systems should have a legal basis and be compliant with any applicable legal regulations. AI fairness requires the fair treatment of sub-populations of users of products involving automation, and ensuring that users are fully aware of the processing to make an informed decision. Transparency means open access to the details of the functionality of an AI product.
\item Security, Data Minimization and Purpose Limitation -  Appropriate security measures should be adopted to decrease the potential for software vulnerabilities and minimize the use of data to reduce the risk of data loss and misuse. Personal data will only be collected and processed to accomplish the specific, explicit, and legitimate purposes
\item Individual Rights - Developing and deploying AI should comply with the individual rights of information, access, rectification, erasure, restriction of processing, data portability, and the right to be informed. 
\end{itemize}

Improperly designed AI systems invite risks of mistreatment of people with certain demographic characteristics. Machine learning bias and fairness has become an active research field. A good survey on recent research results can be found in \cite{mehrabi2021survey}. When used in the HR space, it would violate the equal employment opportunity compliance, which requires treating all people equally when it comes to hiring, promotions, compensation, layoffs, benefits, disciplinary actions and other employment practices. A sample case is that Amazon scraps secret AI recruiting tool that showed bias against women \cite{dastin2018amazon}. An algorithm-driven job advertisement, promoting job opportunities in the science, technology, engineering and math fields, was supposed to be gender neutral, but displayed less to female audience, as the algorithm determined younger women are a prized demographic and are more expensive to show ads to \cite{lambrecht2019algorithmic}. The software, Correctional Offender Management Profiling for Alternative Sanctions (COMPAS), which measures the risk of a person to recommit another crime, is found to be biased against African-Americans \cite{chouldechova2017fair}. In healthcare, a machine learning powered scheduling algorithm is biased against economically disadvantaged groups by assigning them to disfavored time slots, because the algorithm determines those patients have high probability of no-show, given their demographics \cite{samorani2021overbooked}. Machine bias is studied in many other applications like housing \cite{massey2001use}, credit market \cite{fuster2022predictably}.  

Mitigating bias in hiring algorithms requires us to explore how predictive technologies work at each step of the hiring process. Tools used in an early step might be fundamentally different than those used later on. Even the same tools may exhibit disparate behaviors because they were fed with different data sets. Bias can come from different sources \cite{cowgill2018bias}. Historical bias arises when data has historical stereotyping, which
leads to a model that produces harmful
outcomes. Representation bias occurs when the development sample underrepresents some part of the population, and subsequently fails to
generalize well for a subset of the use population. Measurement bias occurs when choosing, collecting, or computing features and labels to use in a prediction problem. Aggregation bias arises when a one-size-fits-all model is used for
data in which there are underlying groups or types of examples that
should be considered differently. Learning bias appears when modeling choices amplify performance disparities across different examples in the data. Evaluation bias occurs when the benchmark data used for a particular task does not represent the use population. 
groups. Deployment bias arises when there is a mismatch between the
problem a model is intended to solve and the way in which it is
actually used. 

Bias also appeared in NLP models. In \cite{bolukbasi2016man}, it shows that  state-of-the-art word embeddings would map ``man" to ``comptuer programmer" and ``woman" to "homemaker". In \cite{rudinger2018gender},  an empirical study shows gender bias in coreference resolution systems. Reference \cite{bordia2019identifying} identifies bias a generated text from a language model based on recurrent neural networks. Reference \cite{may2019measuring} studies bias in sentence embeddings. Reference \cite{font2019equalizing} notices gender bias in machine translation. In \cite{prates2020assessing}, bias is also confirmed in Google translate. Reference \cite{mehrabi2020man} investigates bias in named entity recognition (NER) systems and observe that more female names as opposed to male names are being tagged as non-person entities or not being tagged at all.

\section{Automated Resume Screening System}
A resume screening system's goal is to reduce a large corpus of possible candidates to a manageable amount of resumes. Automated resume screening is used by recruiters and others involved in the hiring process. Larger companies with a greater workforce are more likely to use the system.  Automated screening software helps the hiring managers to quickly review hundreds or even thousands of applications and expedite the hiring process. Existing automated screening software mostly use keyword based methods and categorize resumes based on words. Interestingly, we can find many online tips, teaching how to craft resumes to bypass such keyword based screening software. It is reported in \cite{deshpande2020mitigating} that TF-IDF based resume matching methods possess bias on national origin. keyword based information retrieval has several disadvantages, like low matching accuracy, high computational overhead, etc. Deep learning based information retrieval has attracted a lot of interests from both academia and industry. HR technology companies have now studied the use of deep learning, specifically word embedding, to represent resumes as low-dimensional vectors, which are used for downstream tasks. We term the technology as resume embedding, which is appealing as it transforms a resume into a structured numerical format that is easy to process. However, its involved AI risk has not been investigated. The goal of this study is to examine the technique and address potential concerns if there is any.

\subsection{Resume Embedding}

Resume embedding intends to map a candidate profile to a vector in a relatively low-dimensional space and capture semantics of candidate profiles, such that similar resumes are placed close together in the embedding space. Embedding can create a denser representation of categorical values and maintain some of the implicit relationship information between those values. Embedding can be trained via neural network from data corpus. It has been used in many applications like recommendation systems, computer vision, semantic search, etc.  

A naive resume embedding can be built on existing word embedding. Word embedding is a popular representation of document vocabulary. As opposed to one-hot encoding, another common method for representing categorical variables, which maps a single category to a vector of binary values, word embedding represents words in a low dimensional space and captures context of a word in a document. Popular word embeddings include Word2Vec and GloVe. The Word2Vec algorithm uses a two-layer neural network model to learn word associations from a large corpus of text. It takes as its input a large corpus of text and produces a vector space, typically of several hundred dimensions, with each unique word in the corpus being assigned a corresponding vector in the space. GloVe is another popular word embedding. Its training model is slightly different from Word2Vec, as it stresses the frequency of co-occurrences and can be interpreted as a summary of the training corpus with low dimensionality that reflects co-occurrences. For example, Word2Vec maps ``accountant" to a vector of size 300 as [0.318, -0.234, -0.205, ...] and GloVe converts ``accountant" to a vector of size 100 as [0.20509, -0.45237,  0.26541, ...]. 

To perform resume embedding, we can firstly clean candidate profiles through a set of preprocessing steps, i.e. normalize text, remove Unicode characters like URLs, hashtag, delete stop-words, and perform word stemming and lemmatization. Suppose the cleaned resume $d$ consists of a set of unique terms $\{t_1,...,t_n\}$ with corresponding term frequencies of $\{f_1,..,f_n\}$. Let $W(t_i)$ be the word embedding of term $t_i$. Then, we can derive resume embedding $R(d)$ as:
\begin{equation}
    R(d) = \sum_i f_i W(t_i).
\end{equation}

Resume embeddings can then be used for resume matching. Given a job post $p$, we apply the same preprocessing steps and apply the same word embedding conversion. Denote the resultant embedding as $R(p)$. The matching between resume $d$ and job post $p$ can measured by their cosine similarity:

\begin{equation}
    sim(d, p) = \frac{R(d)R(p)}{|R(d)||R(p)|}.
\end{equation}

\subsection{Resume Embedding Bias and Measure}
Resume embedding indeed significantly improves HR productivity and makes candidate screening efficient. However, it can be vulnerable to stereotypes that are inherited from the training data. Reference \cite{bolukbasi2016man} reports that word embeddings, even training from Google News shows gender stereotypes to a disturbing level. For example, "man" closer to "computer programmer", while "woman" is to "homemaker". Gender bias exhibit in word embedding for many occupation terms. Besides gender bias, many other bias, like nationality origin, age, sexual orientation etc, exist in word embedding as well. Those bias reflect skewed perception in our society. If we apply such word embedding to resume representation, it is expected that the bias will be carried over to automated resume screening. 

This study focuses on the national origin bias in the word embedding-based automated resume screening. According to Equal Employment Opportunity Commission (EEOC), national origin discrimination involves treating people (applicants or employees) unfavorably because they are from a particular country or part of the world, because of their ethnicity or accent, or because they appear to be of a certain ethnic background (even if they are not). The law forbids discrimination when it comes to any aspect of employment, including hiring, firing, pay, job assignments, promotions, layoff, training, fringe benefits, and any other term or condition of employment. Unfortunately, in practice we have seen many reported cases on national origin discrimination. Recruiters may even unconsciously screen out candidate based on certain information on their resume that is associated with some demographic groups. 

National original discrimination occurs in word embedding-based automated resume screening. People from the same national origin tend to use similar vocabularies or have overlaps in interests, due to their cultural background. Those overlap make their resume to exhibit certain pattern, which can be learned and exploited. Our experiments found out that it is possible to recognize a person's national origin to a certain extent, simply based on that person's writing. These days, many intelligent HR tools are able to match job postings with candidate profiles, using AI tools. We found that job postings and profiles, which belong to the same demographic group, can have a higher matching rate. As a result, companies run by a certain ethnicity would only hire/attract people of the same ethical background. The issue in fact has become an obstacle to today's talent acquisition. 

To evaluate and address the issue, we firstly need a fairness measure to determine bias severity. In practice, we can sense that a demographic bias happens if candidates from one demographic group is greatly favored than candidates from another demographic group. We expect that in a "fair" resume screening process, the demographic decomposition of the selected candidates should roughly match that of the entire resume data set where the search is conducted on. For example, if Chinese candidates make up of 1/3 of the resumes collected, the selected Chinese candidates should make up of 1/3 of the selected candidates. To this end, we define a fairness measure as below:
\begin{equation}
    Fairness = \frac{P(selected|d_1)}{P(selected|d_2)}
\end{equation}

where $P(selected|d_1)$ is the chance of a candidate from demographic group $d_1$ being selected by the system, and $d_1$ denotes the demographic group in which a candidate's resume has the lowest chance of being selected, whereas $d_2$ is the demographic group in which a candidate's resume has the highest chance of being selected. A high fairness measure indicates that each demographic groups have about equal chance of being selected by the system, which suggests less or no bias; and low fairness measure can indicate that one demographic group is a lot more favored by the system than another, suggesting more bias.

Another important measure of a resume screening system is its accuracy. An accurate resume screening system means that we can expect the candidates selected by a job posting to be matches to that job posting. For an inaccurate resume screening system, though, a resume being selected by a job posting tells no information of whether it's a match to that job posting. An accuracy measure is proposed as below:

\begin{equation}
Accuracy = P(match|selected).
\end{equation}

\subsection{Bias Mitigation}
Although word embedding converts terms to numerical vectors, those numerical values encode information on national origin. For example, certain food or activities are highly associated with some national origins. Some educational background also imply national origin.  For example, if the word "Shanghai" is identified in a resume or a job posting, it can be reasonably inferred that the resume/job posting is from China. It is known that the appearance of the same term in both a resume and a job posting will result in increased cosine similarity, and thus render the resume more likely to be selected by the system. To resolve this issue, we should remove or decrease the weight of the biased terms to minimize their effect of them. Terms like "Shanghai" or "India" is obviously biased demographically, but other less obvious terms can easily be biased without being recognized as biased terms. 

To mitigate bias, we propose to identify these potentially biased terms and then remove/downgrade their impact on the resume matching algorithm. We use the p-value notion \cite{deshpande2020mitigating} to capture the severity of inherent national origin bias for a term $t$. It is defined as:
\begin{equation}
    p(t)=\frac{P(t|d_{min})}{P(t|d_{max})},
\end{equation}
where $P(t|d)$ denotes the average frequency of a word appearing in one demographic group, and $d_{min}$ is the demographic group where term $t$ is the least likely to be found (lowest $P(t|d)$). $d_{max}$, on the contrary, is the demographic group $d$ that maximizes $P(t|d)$ for term $t$.

 A low $p$ ratio can be an indication of a potentially biased term, because the term is less likely to be find in one demographic than another. Table \ref{tab:pratio of words} shows the corresponding p-ratio value for the terms seen in the word cloud, terms like \textit{management, business}have way higher p-ratio than words like \textit{shanghai, india}. To undermine the effects of these terms, it is intuitive to reduce the weight of the biased terms and promote the weight for those that are considered fair. To this end, a fair resume embedding transformation is introduced to adjust for the term-wise biases
 
 \begin{equation}
    \tilde{R}(d) = \sum_i f_i W(t_i) p(t_i) .
\end{equation}

By introducing the p-ratio value, the above adjusted resume embedding can have less impact from biased words and reduce national origin discrimination. However, it applies adjusted weights to terms proportional to their frequency, which might not be accurate and not yield the optimal result.  To address that, a sigmoid function is used along the p-ratio to generate a weight for each term that is controlled by two parameters: $\lambda$ and $\tau$. The new weighting for each term $t$ is defined as such:
\begin{equation}
    \bar{R}(d) = \sum_i f_i W(t_i) \sigma(\lambda (p(t_i) - \tau ).
\end{equation}

\begin{figure}
    \centering
    \includegraphics[scale=0.5]{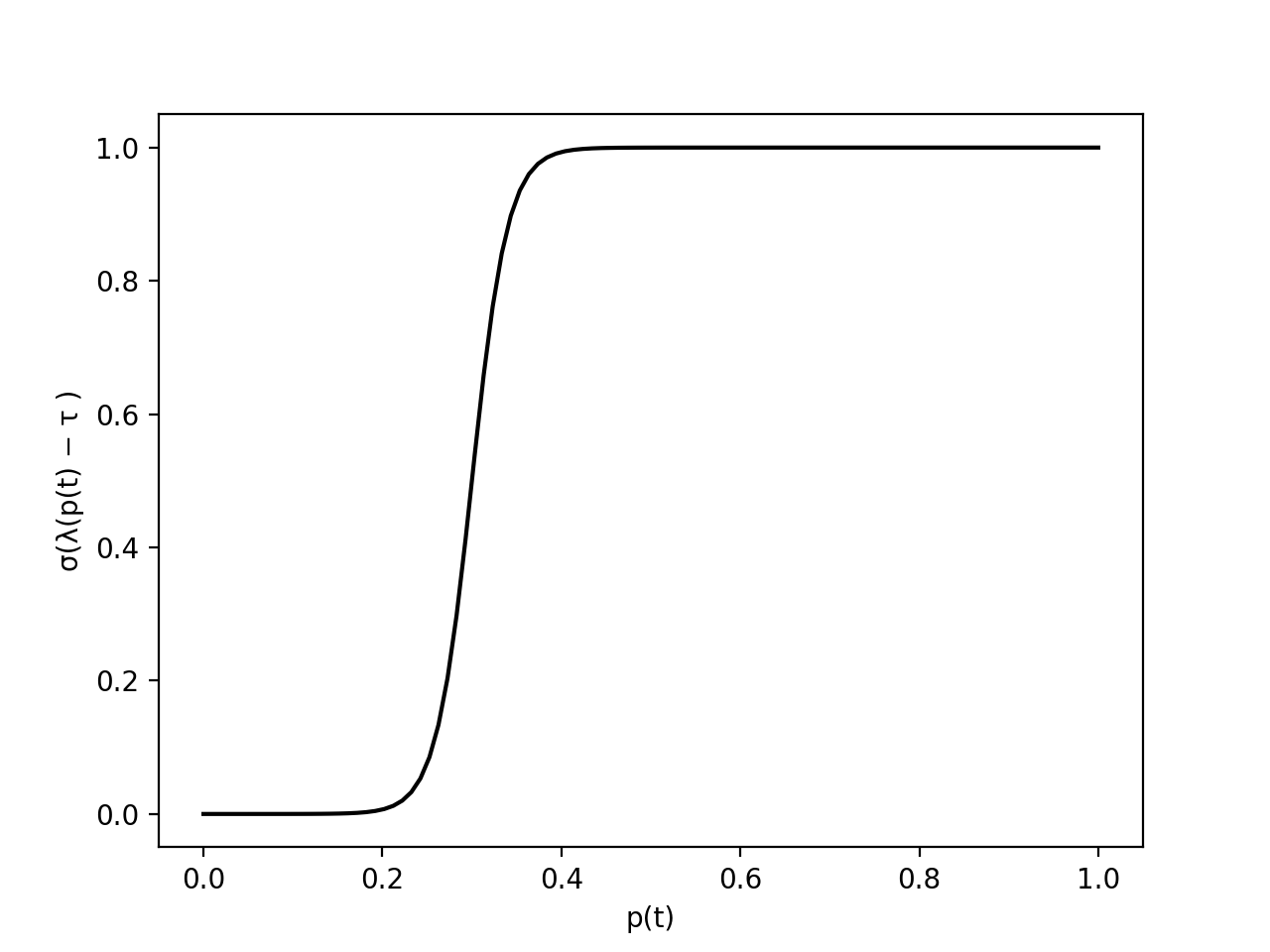}
    \caption{Sigmoid Function with $\lambda=50$ and $\tau=0.3$}
    \label{fig:sigmoid}
\end{figure}

The sigmoid function is a function that convert a linear function to one that is steep near the boundary value, making the output close to 0 for all input values that is below the boundary value and close to 1 for all input values above the boundary values. The two parameters: $\lambda$ and $\tau$ control the steepness of the slope near the boundary value  respectively. The intuition behind this is that if we know certain term is strongly biased, we want to remove it completely, and for the words the we know to be unbiased, we do not reduce its weight even if it might show up a little bit more often in one demographic group than another. The best combination of $\lambda$ and $\tau$ need to be tuned to fit each scenario. To illustrate it, Figure \ref{fig:sigmoid} illustrates the sigmoid function with  $\lambda=50$ and $\tau=0.3$.

\section{Experimental Study}
In this section, we will conduct experiments to examine word embedding-based automated resume screening and its bias towards national origin. In particular, we want to investigate whether a job posting from a certain ethnicity group tends to attract/match resumes with the same ethnicity background, deemed by automated resume screening tools. We will also evaluate our bias mitigation methods. 

\subsection{Dataset and Pre-processing}
\begin{figure}
    \centering
    \includegraphics[height=5cm]{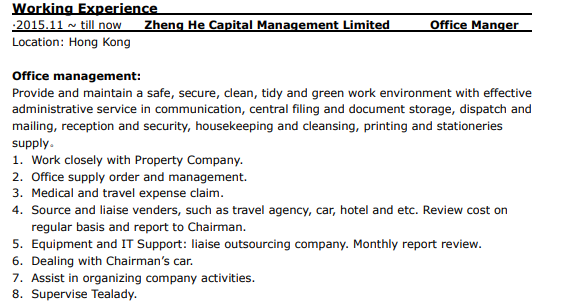}
    \caption{Example of Resume File}
    \label{fig:resume_example}
\end{figure}

We collect two types of datasets, i.e. resumes and job postings. We use the public resume dataset \footnote{https://github.com/JAIJANYANI/Automated-Resume-Screening-System}, which  consists of over 1,000 profiles in PDF or DOCX format. We used libraries including docx2txt\footnote{https://pypi.org/project/docx2txt/} and pdfplumber\footnote{https://github.com/jsvine/pdfplumber} to extract text from the original resume files. The resumes are collected in Singapore, so the demographic/position distribution may match that of Singapore, in which finance, management, and accounting positions divided the dataset evenly. After extracting texts and relevant information from the resumes, 129 resumes from three demographic groups - 64 from China, 31 from India, 34 from Malaysia based on automatic country origin labeling - are selected for further analysis. Out of these, manual check is performed on the labeling to ensure that all labels are accurate. As a result, 28 resumes from India, 48 resumes from China, and 29 resumes from Malaysia (105 resumes in total) are kept. For extracted resumes, we perform standard text cleaning, e.g. removing stop words, word stemming. Figure \ref{fig:resume_example} shows a resume example.

\begin{figure}
     \centering
     \begin{subfigure}[b]{0.33\textwidth}
         \centering
         \includegraphics[width=\textwidth]{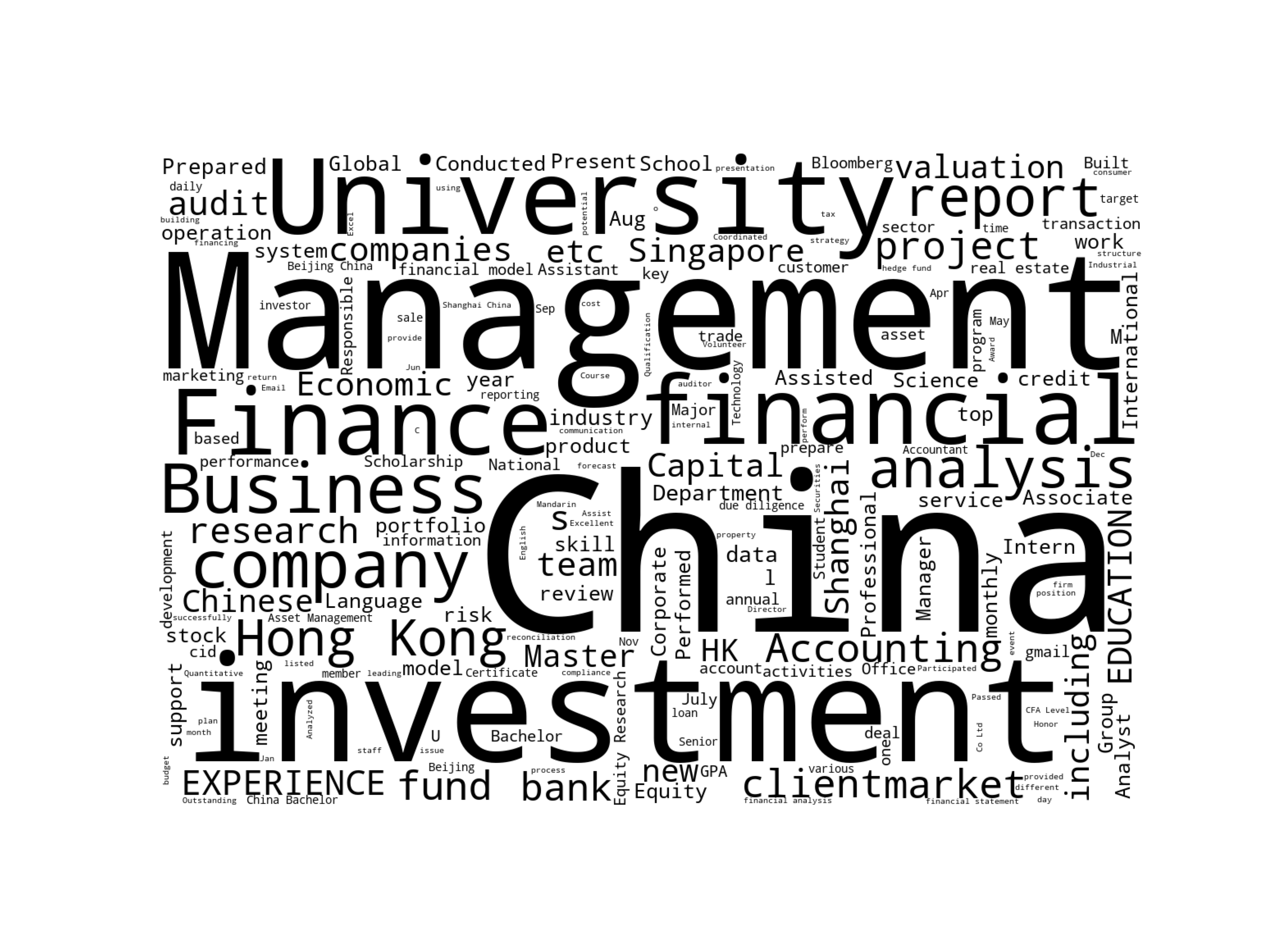}
         \caption{China}
         \label{fig:china wordcloud}
     \end{subfigure}
     \hfill
     \begin{subfigure}[b]{0.33\textwidth}
         \centering
         \includegraphics[width=\textwidth]{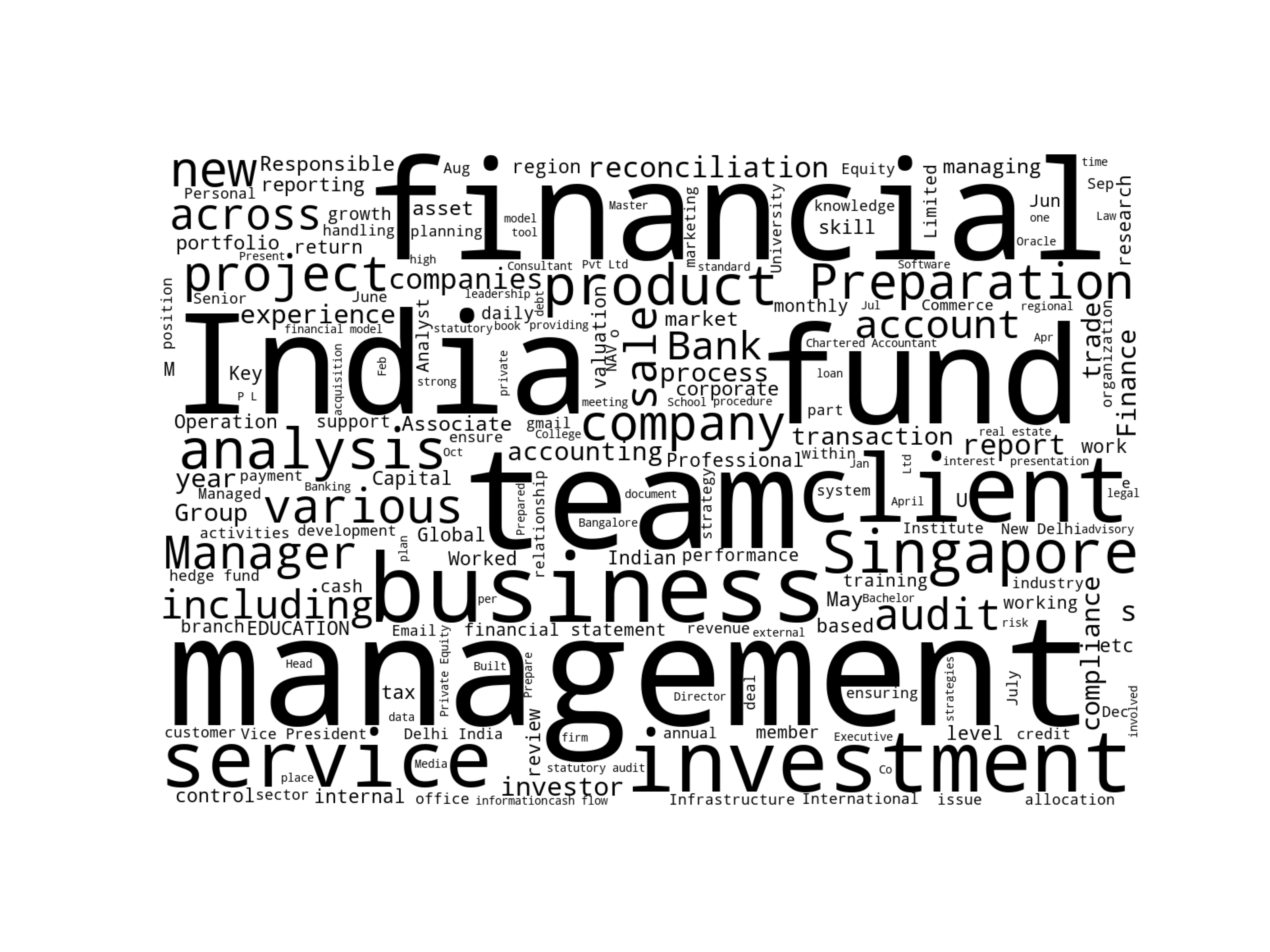}
         \caption{India}
         \label{fig:india wordcloud}
     \end{subfigure}
     \hfill
     \begin{subfigure}[b]{0.33\textwidth}
         \centering
         \includegraphics[width=\textwidth]{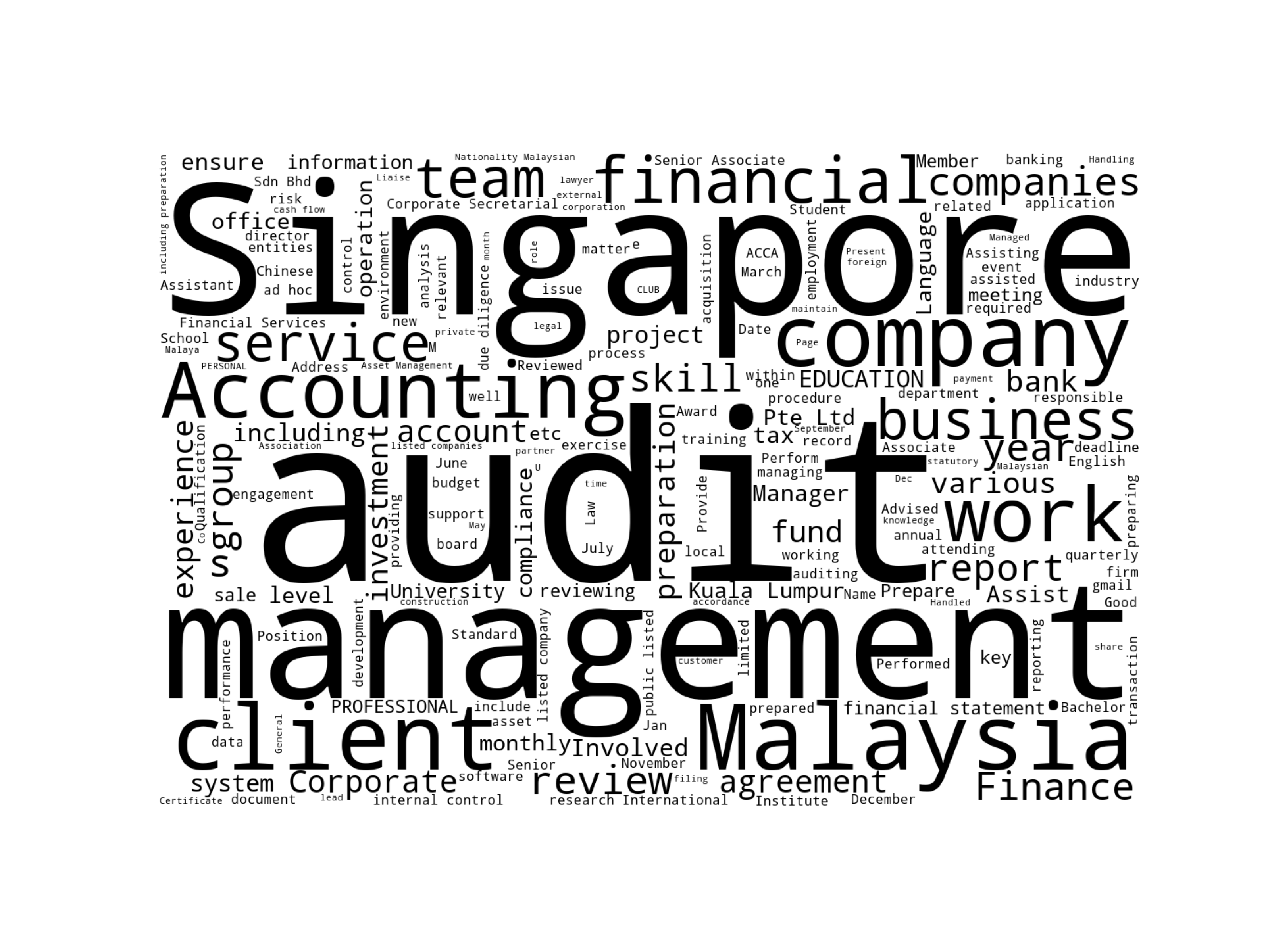}
         \caption{Malaysia}
         \label{fig:five over x}
     \end{subfigure}
        \caption{Word Cloud}
        \label{fig:malaysia wordcloud}
\end{figure}

To get a peek on the candidate resumes from different demographic groups, we generate word cloud for each group. Figures \ref{fig:china wordcloud} - \ref{fig:malaysia wordcloud} provides a visualization of word frequency for each demographic group and clearly shows differences in word uses, although those candidates might have similar experiences and look for the same type of positions. If we purely rely on algorithms to do job matching based on texts, it would inevitably lead to mistakes and bias.

\begin{comment}
\begin{figure}
    \centering
    \includegraphics[height=5cm]{China_wordcloud.png}
    \caption{Word Cloud for Resumes from China}
    \label{fig:china wordcloud}
\end{figure}

\begin{figure}
    \centering
    \includegraphics[height=5cm]{India_wordcloud.png}
    \caption{Word Cloud for Resumes from India}
    \label{fig:india wordcloud}
\end{figure}

\begin{figure}
    \centering
    \includegraphics[height=5cm]{Malaysia_wordcloud.png}
    \caption{Word Cloud for Resumes from Malaysia}
    \label{fig:malaysia wordcloud}
\end{figure}
\end{comment}

For job postings, we collected from popular job posting sites (Indeed.com) to test the resume screening system. While a job posting is usually comprise of parts that states the qualification of the position as well as the parts that advertises the company (the ``What to Expect" part). In terms of screening resume based on a job posting, the ``What to Expect" part contribute little to the searching criteria and will be trimmed when conducting resume screening. For each demographic group that is targeted in this study, three job postings are collected, with each of them resemble one of the common types of role in the dataset: Management Role, Analytic Role, and Accounting Role.

To assess whether the proposed system is accurate, a binary relationship indicating whether a resume matches a job posting is provided between every resume-job posting pair through manual review. This process is done before any further processing took place to ensure the integrity of the experiments. In the ground truth label, 372 out of 945 resume-job posting pairs are labeled as match, so if resumes are randomly selected for any job posting, one should expect about 39\% percent of them to be matched to the job posting. 

\subsection{Results}
The first experiment is to evaluate performance of word embedding based automated resume screening with respect to fairness measure and accuracy. For each ethnic group (i.e. India, China, and Malaysia), we pick 3 job postings. For each job posting, we apply word embedding to generate vector representation. For all candidate profiles, we use the same word embedding to generate resume embedding. For each job posting we retrieve the top 10 profiles in terms of cosine similarity. For each national origin, we report the results averaged over three posting in Table \ref{tab:table1}. The overall accuracy is 0.622, which is much better than random guess. As the first step of a hiring process, 62\% accuracy of automated screening is satisfactory and justifies its use. The fairness measure value is alarming with the overall value of 0.309, whereas the fairness value for India is only 0.165. It suggests that job postings do possess strong signal of national origin and word embedding based resume filtering exhibits evident bias.

\begin{table}
    \centering
    \begin{tabular}{c|c|c}
    \textbf{Country Origin of Job Posting} & \textbf{Fairness Measure} \ & \textbf{Accuracy}\\
    \hline
        India & 0.165 & 0.567\\
        China & 0.292 & 0.633\\
        Malaysia & 0.330 & 0.667\\
        Overall & 0.309 & 0.622\\
    \end{tabular}
    \caption{Word2Vec Fairness Measure for Job Postings from Different Countries}
    \label{tab:table1}
\end{table}

\begin{comment}
\begin{table}
    \centering
    \begin{tabular}{c|c|c}
    \textbf{Country Origin of Job Posting} & \textbf{Fairness Measure} \ & \textbf{Accuracy}\\
    \hline
        India & 0.227 & 0.400\\
        China & 0.493 & 0.467\\
        Malaysia & 0.363 & 0.633\\
        Overall & 0.414 & 0.500\\
    \end{tabular}
    \caption{GloVe Fairness Measure for Job Postings from Different Countries}
    \label{tab:table1}
\end{table}
\end{comment}

We further narrow down candidate profiles and examine accuracy between job positing origin and candidate origin. The results are reported in Table \ref{tab:accuracies_w2v}. For example, for job postings with origin from India, the overall accuracy is 0.567. Among retrieved resumes, the accuracy for candidates from India, China, and Malaysia, is 0.4, 0.8, and 0.467 respectively. As the first glance, it seems no bias issue, as the accuracy for India resumes is lower than China and Malaysia. However, it in fact suggests bias. The accuracy for China resumes is high, because there are fewer resumes being retrieved. On the contrary, more resumes from India being retrieved causes low accuracy for India profiles.

\begin{table}[]
    \centering
    \begin{tabular}{c|llll}
         \textbf{job posting origin} \textbackslash \textbf{candidate origin} &  \textbf{India} & \textbf{China} & \textbf{Malaysia} & \textbf{Overall}\\
         \hline
         \textbf{India} & 0.400 & 0.800 & 0.467 & 0.567\\
         \textbf{China} & 0.714 & 0.667 & 0.545 & 0.633\\
         \textbf{Malaysia} & 0.833 & 0.600 & 0.643 & 0.667\\
         \textbf{Overall} & 0.667 & 0.688 & 0.550 & 0.622\\
    \end{tabular}
    \caption{Accuracy Measure by Job Posting Origin/Candidate Origin Using Word2Vec}
    \label{tab:accuracies_w2v}
\end{table}
\begin{comment}
\begin{table}[]
    \centering
    \begin{tabular}{c|llll}
         \textbf{job posting origin} \textbackslash \textbf{candidate origin} &  \textbf{India} & \textbf{China} & \textbf{Malaysia} & \textbf{Overall}\\
         \hline
         \textbf{India} & 0.500 & 0.454 & 0.333 & 0.400\\
         \textbf{China} & 0.625 & 0.385 & 0.444 & 0.467\\
         \textbf{Malaysia} & 0.833 & 0.800 & 0.429 & 0.633\\
         \textbf{Overall} & 0.667 & 0.529 & 0.395 & 0.500\\
    \end{tabular}
    \caption{Accuracy Measure by Job Posting Origin/Candidate Origin Using GloVe}
    \label{tab:accuracies:glove}
\end{table}
\end{comment}

The second experiment is to evaluate our bias mitigation methods, i.e. p-ratio and sigmoid adjusted p-ratio. The p-ratio method reduces bias by decreasing weights of biased terms. For example, "Shanghai" carries a lot of information national origin. To downgrade its effect, we can multiply its word embedding vector by a value between 0 and 1. The corresponding value is called p-ratio. Its definition can be referred to the previous section. The p-ratio is low (close to 0) for biased terms like "india", "shanghai" and is high (close to 1) for unbiased terms like "finance", "management". Table \ref{tab:pratio of words} reports the p-ratio values for some words. 

\begin{table}[]
    \centering
    \begin{tabular}{c|c}
         \textbf{term} & \textbf{p-ratio} \\
         \hline
         management & 0.966 \\
         finance & 0.900 \\
         audit & 0.574 \\
         business & 0.863 \\
         india & 0.065 \\
         shanghai & 0.144\\
         malaysia & 0.021 \\
    \end{tabular}
    \caption{Sample Terms and Their p-ratio Values}
    \label{tab:pratio of words}
\end{table}

After computing p-ratio values for all terms, we can adjust resume embedding by reducing weights on biased terms and then perform job-resume matching. We carry out the previous experiment with adjusted fair embedding this time. The results are reported in Tables \ref{tab:after fair-embedding w2v} and \ref{tab:accuracy after fair-embedding w2v}. The comparison of fairness measure in Table \ref{tab:after fair-embedding w2v} shows the fair embedding, adjusted by p-ratio values, indeed improves fairness. The overall fairness measure has increased from 0.309 to 0.782, which is a significant jump. The fairness measure for each individual national origin also improves greatly. Particularly, for the Malaysia national origin, it increases from 0.330 to 0.875. Table \ref{tab:accuracy after fair-embedding w2v} reports the accuracy measure for each pair of job posting origin and resume origin. It shows that accuracy values are better than regular word embedding in Table \ref{tab:accuracies_w2v} in general. The results demonstrate that the fair embedding with adjusted p-ratio values performs better than regular word embedding, with respect to both fairness and accuracy. 

\begin{table}[]
    \centering
    \begin{tabular}{c|ll}
         \textbf{Origin of Job Posting} & \textbf{Word-Embedding} & \textbf{Fair-Embedding}  \\
         \hline
         China & 0.292 & 0.429 \\
         India & 0.165 & 0.216 \\
         Malaysia & 0.330 & 0.875 \\
         Overall & 0.309 & 0.782 \\
    \end{tabular}
    \caption{Comparison of Fairness Measure using Word2Vec}
    \label{tab:after fair-embedding w2v}
\end{table}

\begin{table}[]
    \centering
    \begin{tabular}{c|llll}
         \textbf{Job Posting Origin\textbackslash Resume Origin}& \textbf{China} &\textbf{India} &\textbf{Malaysia} &\textbf{Overall}\\
         \hline
         China & 0.643 & 0.500 & 0.667 & 0.600 \\
         India & 0.375 & 1.00 & 0.500 & 0.533 \\
         Malaysia & 0.692 & 0.667 & 0.875 & 0.733 \\
         Overall & 0.600 & 0.652 & 0.625 & 0.622\\
    \end{tabular}
    \caption{Accuracy Measure After Using Fair-Embedding with Word2Vec}
    \label{tab:accuracy after fair-embedding w2v}
\end{table}

\begin{figure}
    \centering
    \includegraphics[height=10cm]{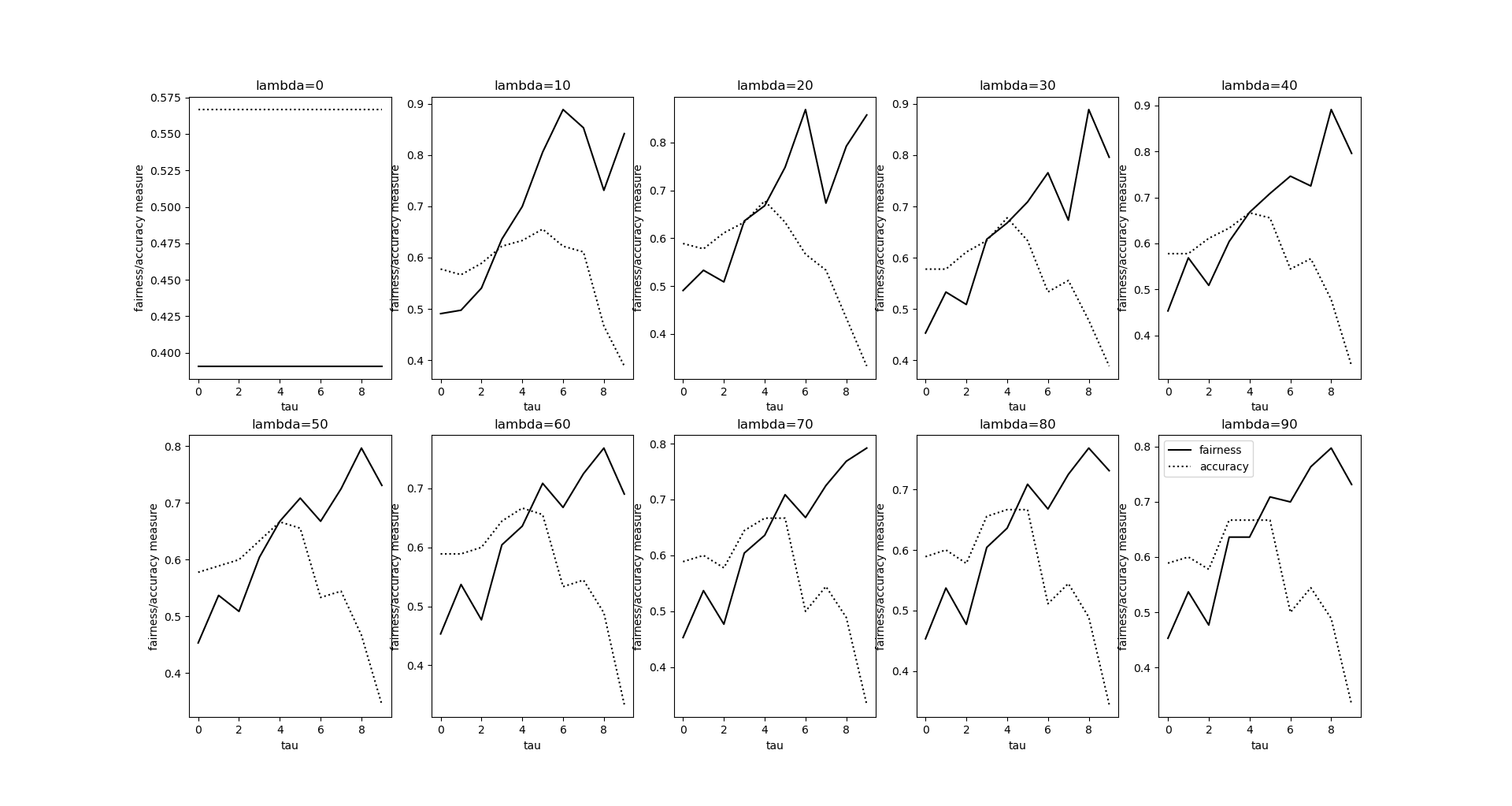}
    \caption{Fairness and Accuracy of different $\lambda$ and $\tau$s}
    \label{fig:tuning}
\end{figure}

The third experiment is to evaluate the adjusted word embedding with sigmoid of p-ratio. As pointed out in the prior section, the p-ratio adjusts term weights proportionally. As such, some terms who occur in most resumes, but do not carry information, would get inappropriate large weights. For example, ``degree" may appear in many resumes. The word alone does to contain much information and should be assigned large weight according to the p-ratio formula. To alleviate the issue,  sigmoid function curves p-ratio, pushes values up when they are above a threshold, and suppresses values down when they are below the threshold. The attempt is to improve the word embedding performance. Recall that the sigmoid function has two parameters, $\lambda$ and $\tau$. They control the steepness of the slope near the boundary value and the boundary value respectively, as shown in \ref{fig:sigmoid}. The intuition behind this is that if we know certain term is strongly biased, we want to remove it completely, and for the words the we know to be not biased, we do not reduce its weight even if it might show up a little bit more often in one demographic group than another. The best combination of $\lambda$ and $\tau$ need to be tuned to fit each scenario. A plot for tuning the parameters can look like this: \ref{fig:tuning}

Following the plot, it can be observed that as $\tau$ increases, the accuracy measure drops. This is intuitive because when we increase the boundary value, terms that are inherently not biased and useful to the screening process are also omitted, causing a decrease in accuracy. There's also a process where accuracy increases with $\tau$, which can possibly be caused by the removal of distraction caused by the biased terms that do not contribute to the resume-job posting analysis. For this case in specific, $\lambda=50$ and $\tau=0.4$ is used as they yield the best performance measure overall.

\begin{table}[]
    \centering
    \begin{tabular}{c|ll}
         \textbf{Origin of Job Posting} & \textbf{Word2Vec With p-Value} & \textbf{Word2Vec  with p-Value and Sigmoid} \\
         \hline
         \textbf{India} & 0.216 & 0.189\\
         \textbf{China} & 0.429 & 0.637\\
         \textbf{Malaysia} & 0.875 & 0.802\\
         \textbf{Overall} & 0.782 & 0.782\\
    \end{tabular}
    \caption{Fairness Measure After Bias Mitigating Techniques}
    \label{tab:fairness after mitigating}
\end{table}

\begin{table}[]
    \centering
    \begin{tabular}{c|llll}
         \textbf{Job Posting Origin\textbackslash Resume Origin}& \textbf{China} &\textbf{India} &\textbf{Malaysia} &\textbf{Overall}\\
         \hline
         China & 0.667 & 0.625 & 0.571 & 0.633 \\
         India & 0.333 & 1.00 & 0.526 & 0.567 \\
         Malaysia & 0.846 & 0.778 & 1.00 & 0.867 \\
         Overall & 0.676 & 0.772 & 0.647 & 0.689\\
    \end{tabular}
    \caption{Accuracy Measure After Using Fair-Embedding-Sigmoid Word2Vec}
    \label{tab:accuracy after fair-embedding-sigmoid w2v}
\end{table}

We apply the fair embedding with sigmoid of p-ratio to the same dataset. The results are reported in Table \ref{tab:fairness after mitigating}, \ref{tab:accuracy after fair-embedding-sigmoid w2v}. We see that while the fairness measure doesn't change much at this $\lambda$ and $\tau$ values, the accuracy measure experienced an increase as a result of our tuning strategy. The sigmoid function provides a way to find a sweet spot between the fairness-accuracy trade-offs.

\begin{figure}
    \centering
    \includegraphics[height=7cm]{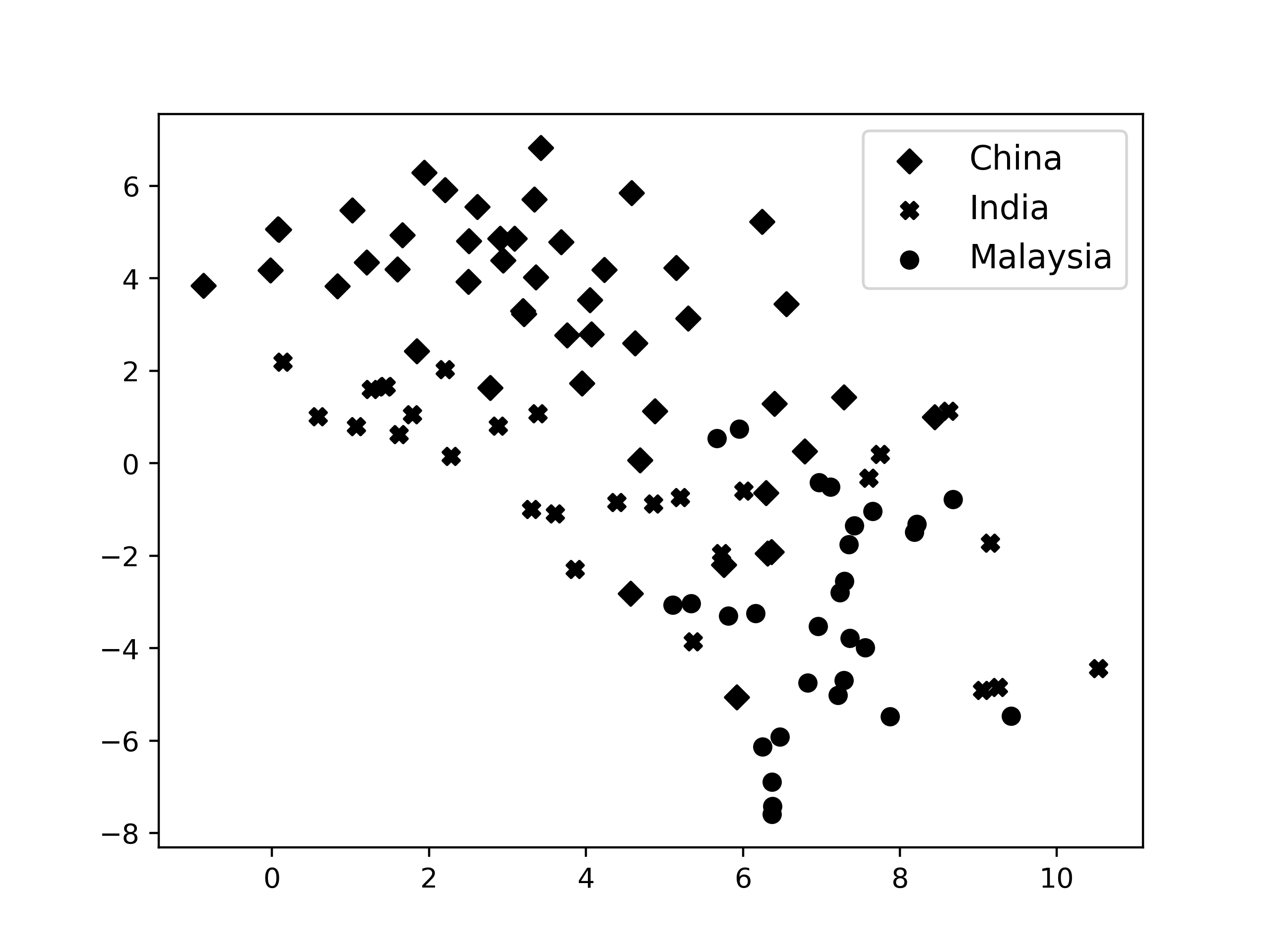}
    \caption{T-SNE Before Applying p-Value fairness}
    \label{fig:tsne before}
\end{figure}

\begin{figure}
    \centering
    \includegraphics[height=7cm]{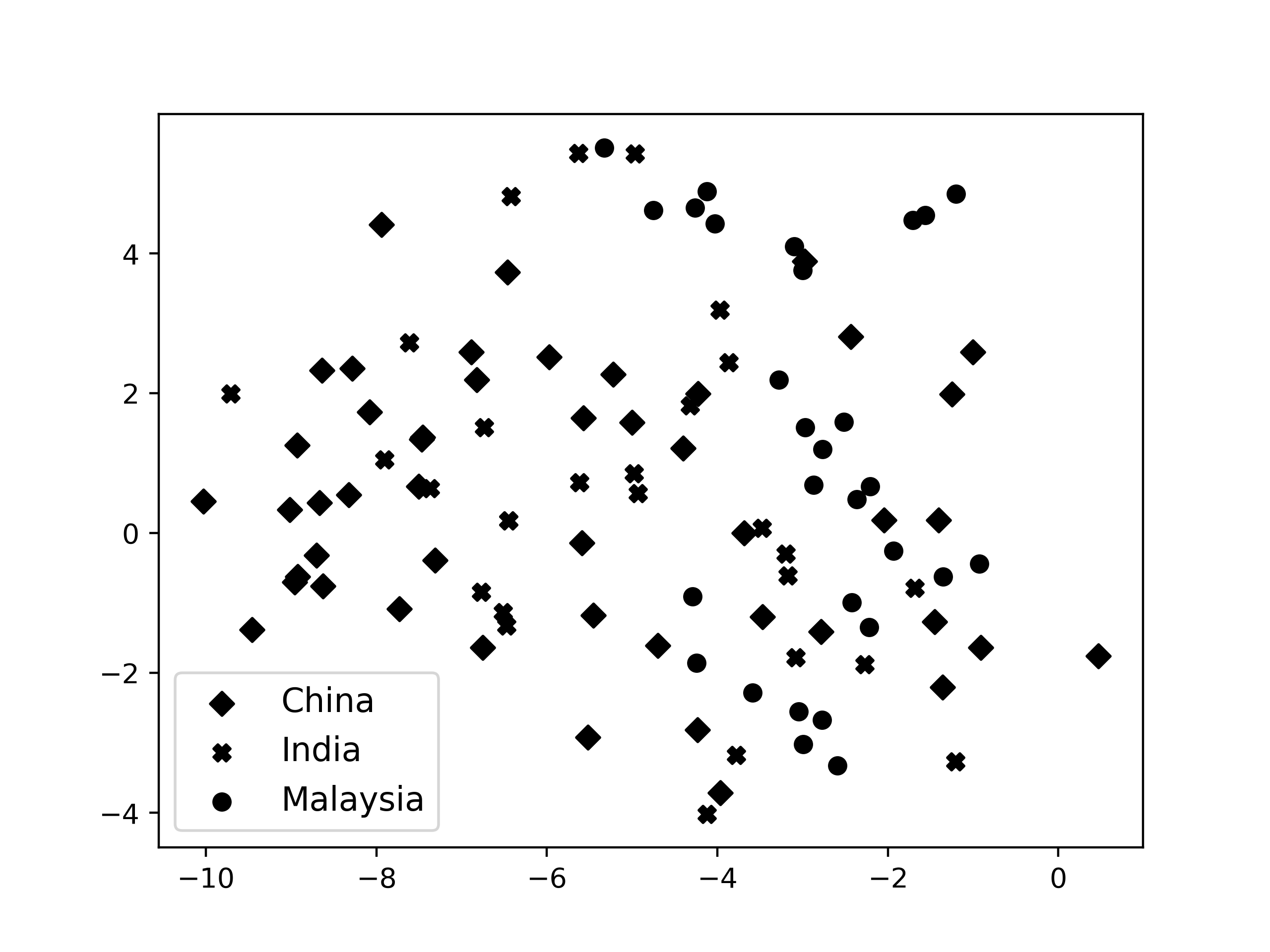}
    \caption{T-SNE After Applying p-Value fairness}
    \label{fig:tsne fairtf}
\end{figure}

\begin{figure}
    \centering
    \includegraphics[height=7cm]{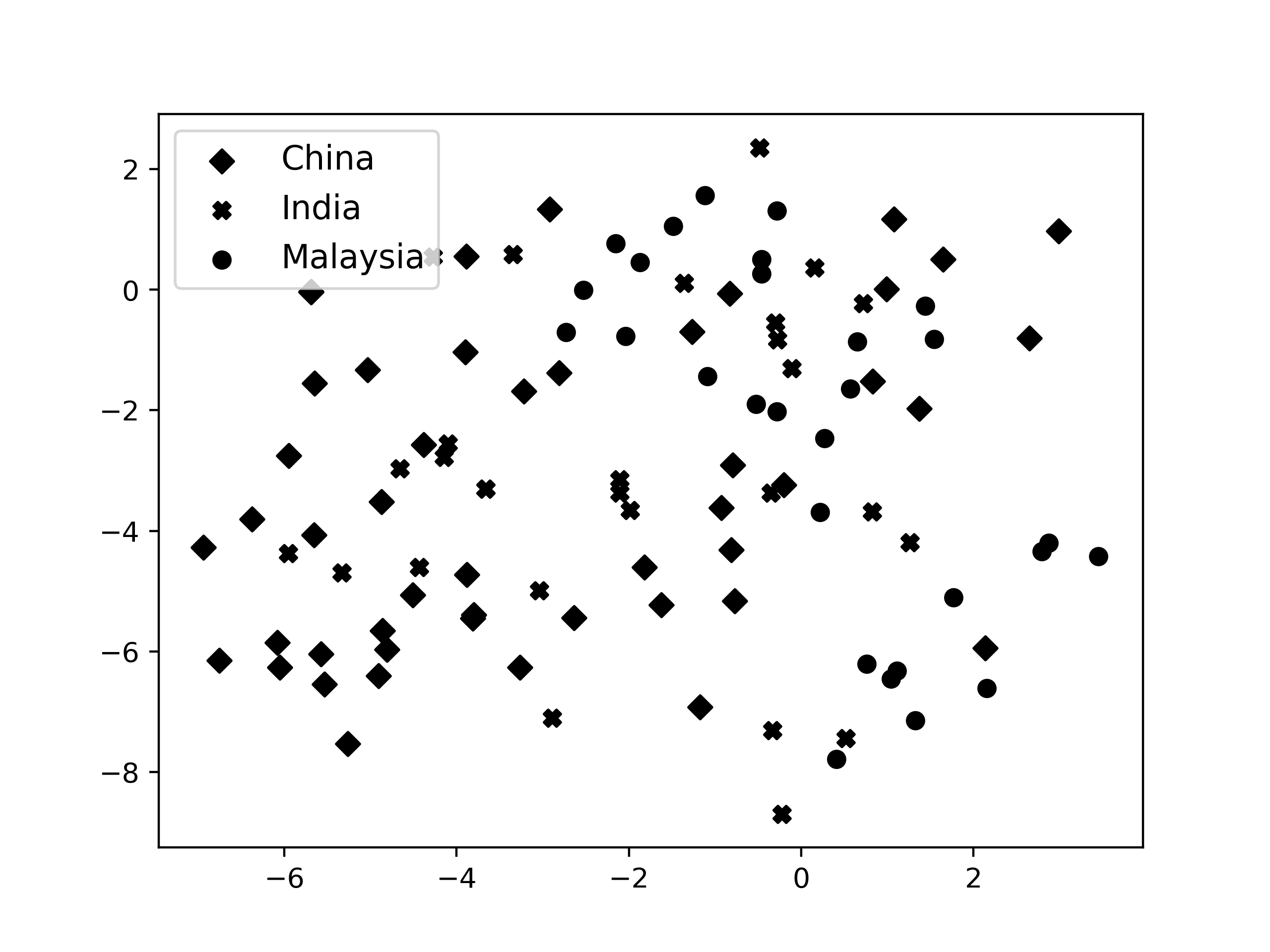}
    \caption{T-SNE After Applying p-Value fairness with sigmoid}
    \label{fig:tsne sigmoid}
\end{figure}

To visualize the effects of different resume embedding methods, we utilize T-Distributed Stochastic Neighbor Embedding (T-SNE) plots \cite{van2008visualizing}, which is a popular statistical method for visualizing high-dimensional data by giving each datapoint a location in a two or three-dimensional map. The visualizations are plotted in Figure \ref{fig:tsne before}, \ref{fig:tsne fairtf}, \ref{fig:tsne sigmoid} provide a visual representation of how fair embedding with p-ratio and fair embedding with sigmoid of p-ratio are able to mitigating the demographically separated clusters in the embedded resume space. The clustering boundary is clear and well defined in \ref{fig:tsne before}, and the clusters start to mix together in \ref{fig:tsne fairtf} and \ref{fig:tsne sigmoid}, as a result of removal or lessening the weight of the intra-cluster similarities.

\begin{figure}
    \centering
    \includegraphics[height=6cm]{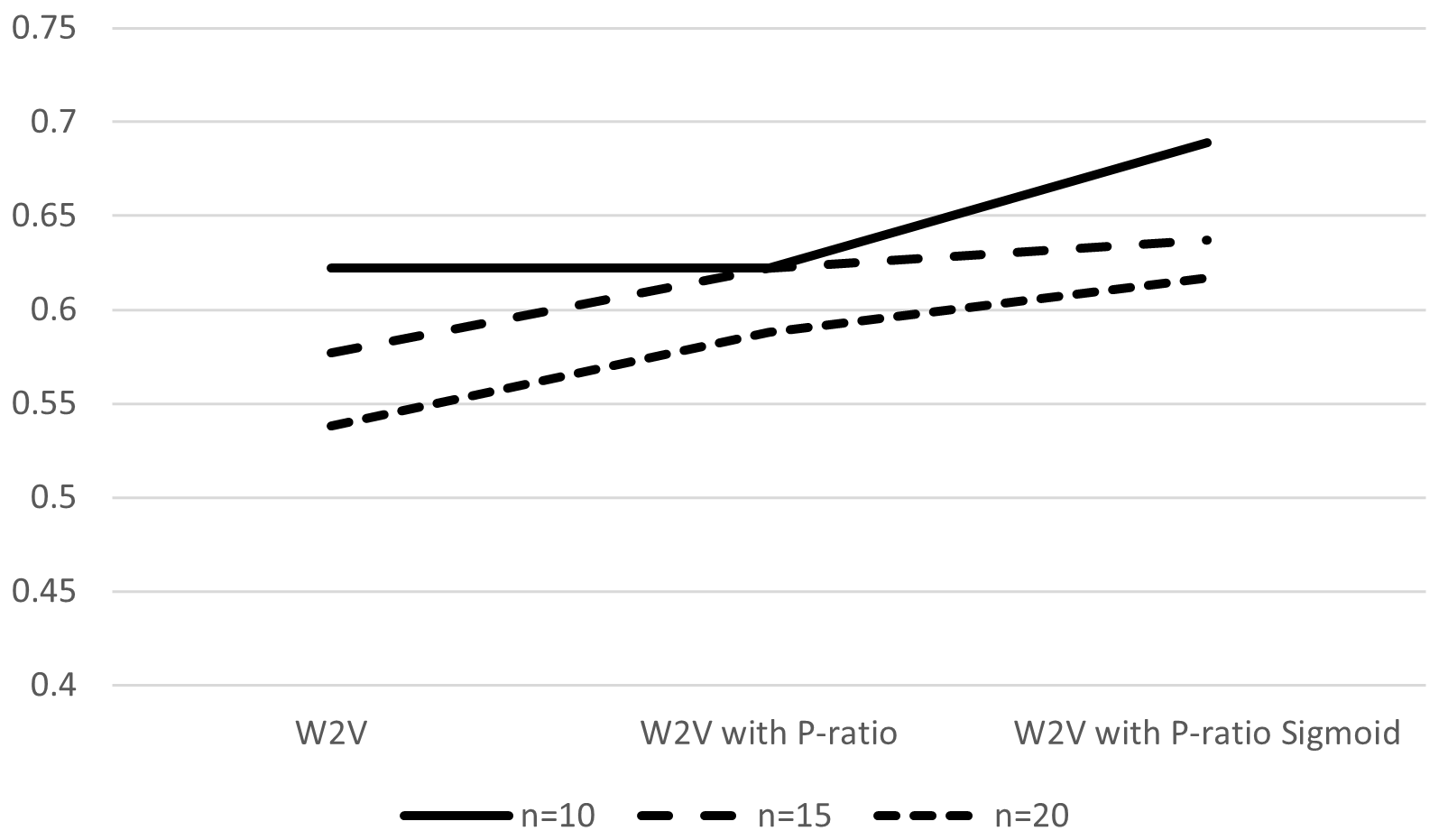}
    \caption{Accuracy Measure with Different N Values}
    \label{fig:accuracy_varying_n}
\end{figure}

\begin{figure}
    \centering
    \includegraphics[height=6cm]{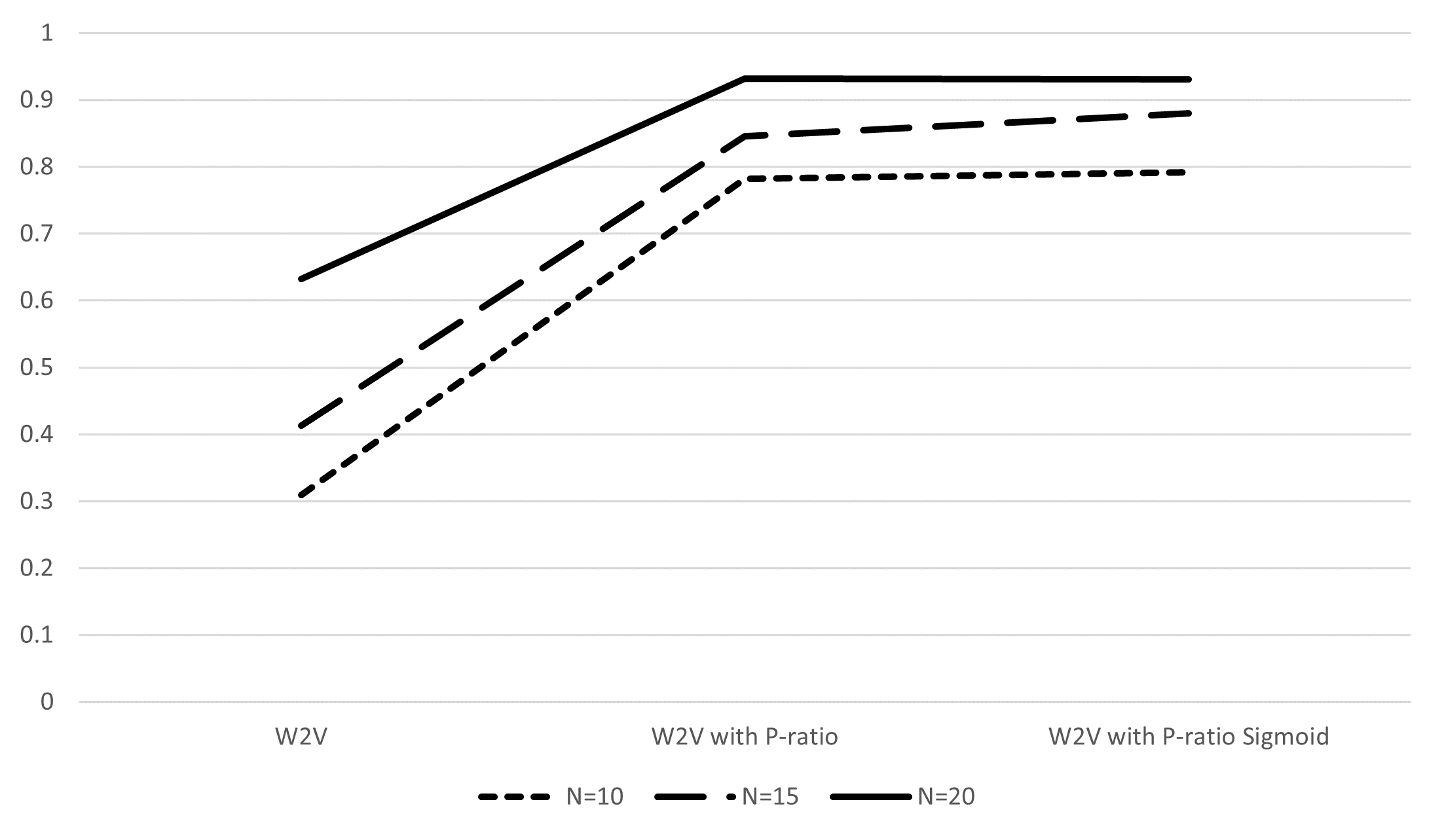}
    \caption{Fairness Measure with Different N Values}
    \label{fig:fairness_varying_n}
\end{figure}

Then, we conduct another experiment to see how the number of top matched resumes affect the result. In the exemplary result system, 10 resumes at a time is selected from the resume pool for each query job descriptions (N=10). In reality, however, the user should be able to customize how many resume should be screened. The choice of N can affect the initial accuracy/fairness measure dramatically if not chosen carefully: for example, if N is too small, there is a higher chance to get a fairness measure of 0 is none of the resumes from a demographic is selected. If N is too big that it covers almost all of the resume, then the fairness measure will be higher, which can underestimate the bias in the system. The bias mitigating techniques, however work regardless of the choice of N (\ref{fig:accuracy_varying_n}, \ref{fig:fairness_varying_n}, \ref{tab:accuracy before fair-embedding w2v n=15}, \ref{tab:accuracy after fair-embedding w2v n=15}, \ref{tab:accuracy after fair-embedding-sigmoid w2v n=15}). It turns out that, regardless of the N value to choose, the bias mitigating techniques are always able increase (or stays the same) both performance measures.

\begin{table}[]
    \centering
    \begin{tabular}{c|llll}
         \textbf{Job Posting Origin\textbackslash Resume Origin}& \textbf{China} &\textbf{India} &\textbf{Malaysia} &\textbf{Overall}\\
         \hline
         China & 0.571 & 0.588 & 0.571 & 0.578 \\
         India & 0.333 & 0.769 & 0.478 & 0.533 \\
         Malaysia & 0.600 & 0.714 & 0.571 & 0.622 \\
         Overall & 0.515 & 0.682 & 0.534 & 0.578\\
    \end{tabular}
    \caption{Accuracy Measure Before Fairness-Embedding Word2Vec n=15}
    \label{tab:accuracy before fair-embedding w2v n=15}
\end{table}

\begin{table}[]
    \centering
    \begin{tabular}{c|llll}
         \textbf{Job Posting Origin\textbackslash Resume Origin}& \textbf{China} &\textbf{India} &\textbf{Malaysia} &\textbf{Overall}\\
         \hline
         China & 0.640 & 0.500 & 0.625 & 0.600 \\
         India & 0.500 & 0.857 & 0.462 & 0.533 \\
         Malaysia & 0.632 & 0.750 & 0.857 & 0.733 \\
         Overall & 0.607 & 0.677 & 0.604 & 0.622\\
    \end{tabular}
    \caption{Accuracy Measure After Fairness-Embedding Word2Vec n=15}
    \label{tab:accuracy after fair-embedding w2v n=15}
\end{table}

\begin{table}[]
    \centering
    \begin{tabular}{c|llll}
         \textbf{Job Posting Origin\textbackslash Resume Origin}& \textbf{China} &\textbf{India} &\textbf{Malaysia} &\textbf{Overall}\\
         \hline
         China & 0.667 & 0.400 & 0.625 & 0.600 \\
         India & 0.462 & 0.875 & 0.500 & 0.556 \\
         Malaysia & 0.565 & 0.700 & 1.00 & 0.711 \\
         Overall & 0.587 & 0.643 & 0.660 & 0.622\\
    \end{tabular}
    \caption{Accuracy Measure After Fairness-Embedding-Sigmoid W2V n=15}
    \label{tab:accuracy after fair-embedding-sigmoid w2v n=15}
\end{table}

% \subsection{Different Word Embedding}

The Google News Word2Vec model is used in this experiment. Because this model is general purpose rather than specialized, may not represent the best performance that can be achieved by a word-embedding based resume screening system. The bias mitigating techniques presented, however, works on other word-embedding model as well as the Google News Word2Vec. As shown in the table \ref{tab:fairness glove} and \ref{tab:accuracy glove}, the GloVe word-embedding performs inferior to the Google News Word2Vec model at both measures initially. The mitigating techniques, including P-ratio and P-ratio with sigmoid, however, still managed to increase its fairness measure and accuracy measures. 

\begin{table}[]
    \centering
    \begin{tabular}{c|lll}
         \textbf{Job Posting Origin} & \textbf{GloVe} & \textbf{GloVe with P-ratio} &\textbf{GloVe with P-ratio Sigmoid}\\
         \hline
         India & 0.250 & 0.232 & 0.252 \\
         China & 0.371 & 0.713 & 0.473 \\
         Malaysia & 0.302 & 0.544 & 0.891 \\
         Overall & 0.428 & 0.640 & 0.760 \\
    \end{tabular}
    \caption{Fairness Measure Utilizing Each Techniques with GloVe}
    \label{tab:fairness glove}
\end{table}

\begin{table}[]
    \centering
    \begin{tabular}{c|lll}
         \textbf{Job Posting Origin} & \textbf{GloVe} & \textbf{GloVe with P-ratio} &\textbf{GloVe with P-ratio Sigmoid}\\
         \hline
         India & 0.400 & 0.467 & 0.567 \\
         China & 0.467 & 0.433 & 0.567 \\
         Malaysia & 0.644 & 0.700 & 0.800 \\
         Overall & 0.500 & 0.533 & 0.644 \\
    \end{tabular}
    \caption{Accuracy Measure Utilizing Each Techniques with GloVe}
    \label{tab:accuracy glove}
\end{table}

\section{Conclusion and Discussion}
AI has been viewed as a technological revolution and provided unprecedented opportunities to our society. However, opportunities also come with challenges and even harm. If we cannot well understand AI technologies and their inherent fairness/bias issues, AI can cause great damage and create ethics and societal issues on a large scale. This study is following the active research stream on AI bias in automated resume filtering, focusing on the application of word embedding in candidate profile matching. Our study finds out that the word embedding (learned from a large corpus with neural networks) based job-resume matching algorithm can carry national origin bias, which reflects biases of individual terms. Using such systems, national origin bias would be amplified in work forces and can invite lawsuits. To address the issue from the algorithm perspective, we introduce several improved algorithms, inspired by existing research results. The key idea is to adjust weights on individual terms when generating resume vectors, by reducing the impact of terms with strong national origin bias. Extensive experiments are conducted to evaluate the proposed algorithms. The results suggest that our algorithms have great performance with respect to the fairness measure and accuracy. 

Since the weights of each term in the proposed algorithms are learned from the training data, the algorithm may require a medium to large scale of labeled data when applied in the real world. When it comes to tiny-scale tasks like reviewing a single resume, other bias-mitigating techniques might be turned to. While the algorithm is supposed to reduce biases for all demographic groups divided along the protected attributes (gender, race, religion, etc), extra work might need to be done to obtain labeled data set for some of the attributes. 

Algorithmic fairness in HR technologies is a broad subject. There are a lot of studies to be done. For example, it is worth investigating how to adjust word vectors directly along some subspace, such as national origin, to remove inherent bias, as opposed to adjusting the weights of terms. Such a solution would be more elegant and useful. Some protected attributes, like sexual orientation and disability status, might be difficult to infer from resume text. However, their bias does exist in the AI world. When there is no training data to reflect those attributes' information, how to devise algorithms to remove such bias? In addition to resume filtering, bias appears in other phases of AI-assisted hiring as well. Exploring those issues will be left as our future work.

%Bibliography
\bibliographystyle{unsrt}  
\bibliography{references}

\end{document}